\documentclass{article} 
\usepackage{iclr2025_conference,times}

\usepackage{amsmath,amsfonts,bm}

\def\eqref#1{equation~\ref{#1}}

\def\1{\bm{1}}

\DeclareMathAlphabet{\mathsfit}{\encodingdefault}{\sfdefault}{m}{sl}
\SetMathAlphabet{\mathsfit}{bold}{\encodingdefault}{\sfdefault}{bx}{n}

\iclrfinalcopy
\usepackage{hyperref}
\usepackage{url}
\usepackage{graphicx}
\usepackage{float}
\usepackage{subcaption}
\usepackage{booktabs}

\title{The role of positional encodings in the ARC benchmark}

\author{
  Guilherme H. Bandeira Costa \\
  INESC-ID, NeuralShift \\
  \texttt{guilherme.costa@neuralshift.ai} \\
  \And
  Miguel Freire \\
  NeuralShift \\
  \texttt{miguel.freire@neuralshift.ai} \\
  \AND
  Arlindo L. Oliveira \\
  INESC-ID \\
  \texttt{arlindo.oliveira@tecnico.ulisboa.pt} \\
}

\begin{document}

\maketitle
\pagestyle{plain}

\begin{abstract}
The Abstraction and Reasoning Corpus challenges AI systems to perform abstract reasoning with minimal training data, a task intuitive for humans but demanding for machine learning models. Using CodeT5+ as a case study, we demonstrate how limitations in positional encoding hinder reasoning and impact performance. This work further examines the role of positional encoding across transformer architectures, highlighting its critical influence on models of varying sizes and configurations. Comparing several strategies, we find that while 2D positional encoding and Rotary Position Embedding offer competitive performance, 2D encoding excels in data-constrained scenarios, emphasizing its effectiveness for ARC tasks. 
\end{abstract}

\section{Introduction}

\begin{figure}[ht]
  \centering
  \includegraphics[width=0.9\linewidth]{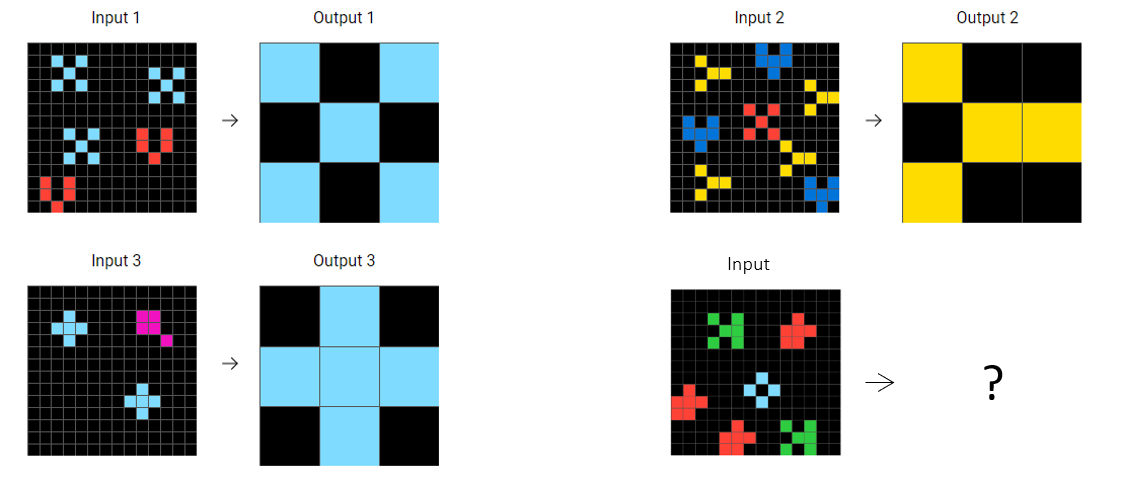}
     \caption {Example of an ARC task. All images that include both input and output are demonstration examples, the image that only includes input is the test example. In this particular task, the rule is that the output grid corresponds to representing the most common object, differentiated by different colors, in the input grid.}
  \label{Example_ARC}
\end{figure}

With the recent competition surrounding the Abstraction and Reasoning Corpus (ARC) challenge \citep{ARC2024}, more participants have attempted to tackle its unique tasks, pushing the boundaries of what Artificial Intelligence (AI) systems can achieve. Designed by \citet{OntheMeasureofIntelligence}, ARC serves as a benchmark for evaluating an AI system’s ability to perform abstract reasoning by requiring models to transform input grids into output grids based on minimal training examples, each illustrating a specific reasoning process (see Figure \ref{Example_ARC}). Although these tasks are intuitive for humans, who can easily generalize from a few examples, they pose significant challenges for machine learning algorithms \citep{Comparing}, which excel at curve fitting but struggle with reverse engineering diverse implicit rules. 

Recent advances, such as OpenAI's O3 model, have shown promise in the ARC benchmark \citep{Chollet_O3}. However, these successes come at an immense computational cost, with O3 spending thousands of dollars to solve a single task. We posit that such performance could be achieved more efficiently by reconsidering key components of transformer-based models \citep{attentionisallyouneed}, particularly their positional encoding mechanism. If positional encoding was designed to better align with ARC’s spatial reasoning requirements, similar performance might be attainable with far fewer resources.

In this work, we investigate the role of positional encoding in influencing the reasoning of a model in ARC tasks. Using a simple example, we show how CodeT5+ \citep{CODET5+}, a Large Language Model (LLM), fails due to limitations in its positional encoding mechanism. We further evaluate the impact of different positional encodings across various transformer architectures, demonstrating that effective encoding significantly enhances performance regardless of model size or structure. This is achieved by comparing the original positional encoding proposed in \citet{attentionisallyouneed} and its 2D extension \citep{2DPE}, which incorporates the x and y positions of a token to calculate its encoding. Finally, we compare several positional encoding strategies, including 2D encodings, Rotary Position Embedding (RoPE) \citep{RoPE}, and Learned Embeddings, a trainable alternative to fixed encodings first introduced in \citet{attentionisallyouneed}. Our results indicate that in data-constrained scenarios like ARC, 2D positional encoding outperforms other approaches, while RoPE demonstrates a slight performance advantage in high-data settings, offering a viable alternative when substantial training data is available.

\section{Background}

\textbf{Positional Encoding.} Incorporating positional information is essential for transformer models, enabling them to capture order and spatial relationships that would otherwise be missing in their inherently order-agnostic architectures. The original sinusoidal encoding introduced by \citet{attentionisallyouneed} uses fixed sine and cosine functions to provide smooth transitions across sequence lengths. Learned embeddings, a trainable alternative, adapt positional representations during training but often yield limited improvements.

For tasks involving grids, such as ARC, 2D sinusoidal encodings \citep{2DPE} extend the original concept by independently encoding horizontal and vertical positions, effectively capturing 2D spatial relationships. Relative positional encodings \citep{RelativePE} further enhance this by encoding token-to-token distances rather than absolute positions, excelling in sequences but proving less effective in grid-based structures. RoPE \citep{RoPE} encodes relative positions by applying rotational transformations to query and key vectors, making it particularly effective in capturing long-range dependencies.
Tailored approaches such as Abacus Embeddings \citep{AbacusEmbeddings} demonstrate that task-specific encodings can surpass general-purpose methods in domains requiring precise positional awareness.

\textbf{Related Work.} Evidence provided by \citet{2DVITARC} demonstrates that a two-dimensional visual representation significantly enhances the reasoning performance of vision transformers, with positional information further improving their capabilities in spatially complex tasks. However, to the best of our knowledge, no specific experiments have evaluated the impact of positional encoding on ARC using the vanilla transformer architecture.

\section{How does positional encoding impact LLM reasoning}
\begin{figure}[H]
  \centering
  \includegraphics[width=0.7\linewidth]{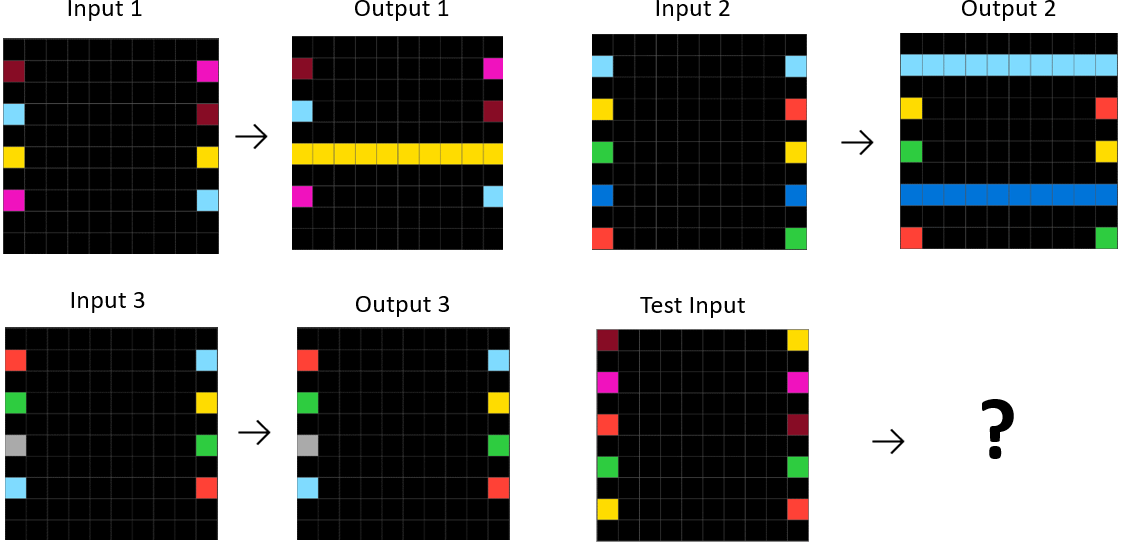}
     \caption[Chosen task for the CodeT5+ experiments]{The selected task is derived from the original ARC dataset where the objective is to connect rows where the pixels at both ends share the same color. Importantly, if all examples are rotated by 90 degrees, the goal of the task remains unchanged, except that the connections would then occur across the columns instead of the rows.}
  \label{Line_Drawing_Task}
\end{figure}

\begin{figure}[H]
  \centering
  \includegraphics[width=0.7\linewidth]{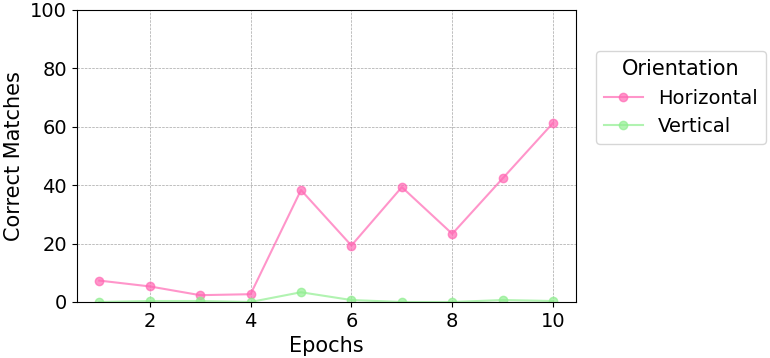}
     \caption{Average correct matches across 3 distinct batch sizes when trained and tested for horizontal (pink) versus vertical (green) examples using the default CodeT5+ model.}
  \label{HorizontalvsVertical}
\end{figure}

Positional encoding is a fundamental component of transformer-based models such as CodeT5+, enabling them to process sequential data and understand the relationships between tokens. To demonstrate its influence, we utilize a simple ARC task whose goal involves connecting lines between pixels. Although this task presents the same reasoning challenges for humans regardless of whether it is presented in a horizontal or vertical orientation (see Figure \ref{Line_Drawing_Task}), positional encoding can significantly influence the performance of an LLM in the two distinct views.
To demonstrate this, we used an ARC generator \citep{MichaelHodel}  to create horizontal examples for the ARC task and rotated them 90 degrees to generate vertical examples, ensuring that they have the same level of difficulty. Using these examples, we trained two CodeT5+ models: one exclusively on horizontal examples and the other on vertical examples. Despite the identical structure of the task, the model performed considerably better on horizontal examples (see Figure \ref{HorizontalvsVertical}). Given that the attention mechanism in transformer-based models is designed to attend to all positions, this disparity suggests that the positional encoding used by CodeT5+ plays a critical role in shaping its ability to handle different spatial relationships.

\begin{figure}[H]
  \centering
  \includegraphics[width=0.6\linewidth]{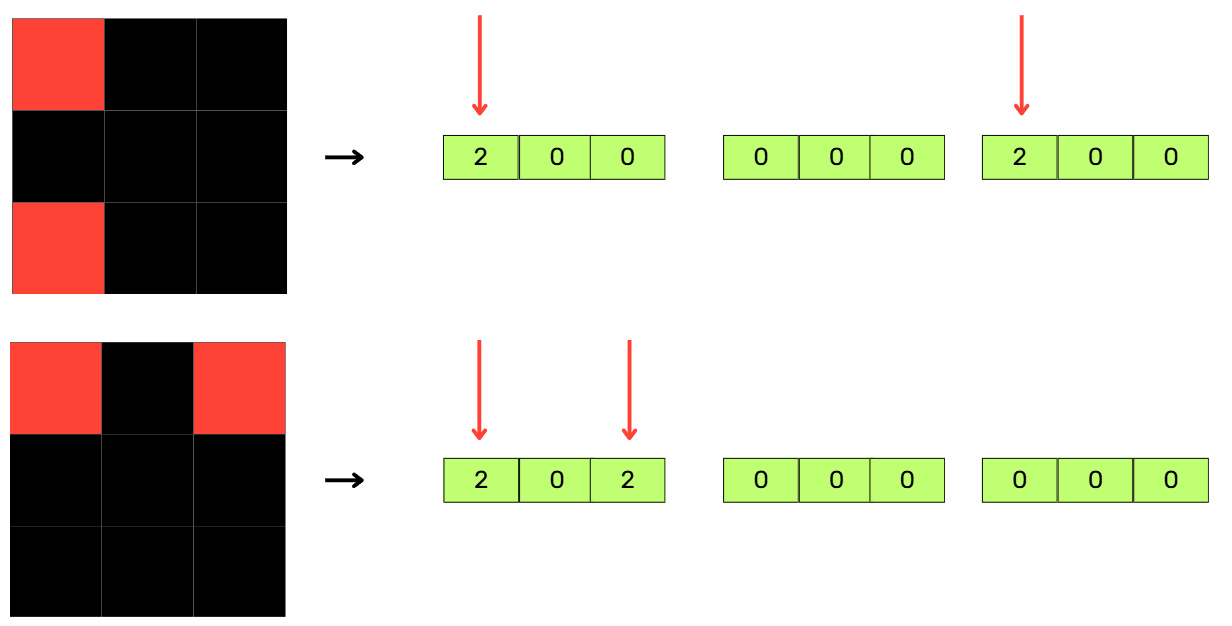}
     \caption[Horizontal vs Vertical Examples Token Distance]{Visual illustration of token creation for vertical (top) and horizontal (bottom) examples. Tokens are generated linearly, with the ``goal tokens'' (representing the red color) appearing closer together in horizontal arrangements and farther apart in vertical ones.}
  \label{HorizontalvsVertical Tokens}
\end{figure}

In fact, upon analyzing the CodeT5+ positional encoding mechanism, we find that it employs relative positional encoding, which prioritizes tokens that are closer together in the sequence. This explains the observed disparity: In horizontal examples, the relevant tokens are naturally closer together, making it easier for the model to establish the necessary relationships. In contrast, in vertical examples, the relevant tokens are farther apart, spanning multiple rows, making it more difficult for the model to capture these relationships (see Figure \ref{HorizontalvsVertical Tokens}).

\begin{figure}[H]
  \centering
  \includegraphics[width=0.7\linewidth]{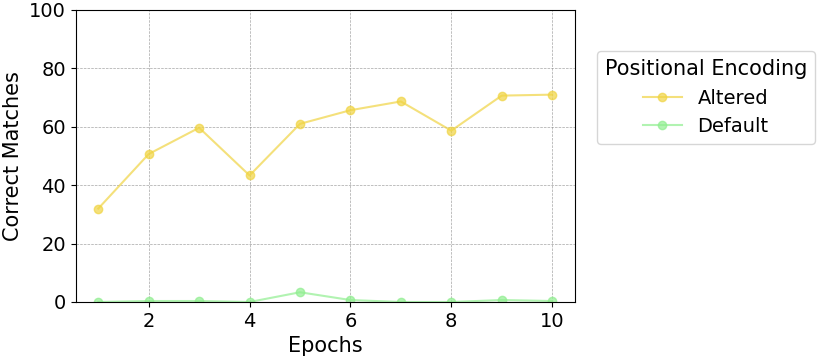}
     \caption{Average correct matches across 3 distinct batch sizes when trained and tested for vertical examples with the original PE (green) versus an altered PE which prioritizes relevant tokens (yellow) using the default CodeT5+ model.}
  \label{DefaultvsAlteredPE}
\end{figure}

To address this limitation, we modified the positional encoding to prioritize relationships specifically between the "goal tokens" in vertical examples. With this adjustment, the model’s performance on vertical tasks improved significantly (see Figure \ref{DefaultvsAlteredPE}), showing the clear impact that positional encoding has on the model’s understanding and reasoning.

\section{Experiments}
Since CodeT5+ was pre-trained with a deeply integrated positional encoding mechanism, modifying it to effectively investigate the impact of positional encoding posed significant challenges. To address this, we develop several custom transformer models from scratch, enabling a systematic evaluation of how different positional encoding strategies influence performance across various model architectures.

\subsection{Experimental setup} We trained the models on 100,000 examples for each of 10 distinct ARC tasks independently to ensure that the results were not biased towards any single task. The evaluation is performed every two epochs, following the method proposed in \citet{OntheMeasureofIntelligence}, where an answer is considered correct only if all pixels are in the correct position. The results are presented as aggregated averages across all tasks.

The experiments were designed to systematically assess the impact of different positional encoding strategies and input formats. First, we compared various model architectures using the original positional encoding proposed in \citet{attentionisallyouneed}, which we refer to as 1D, with two distinct tokenization formats: raw and bracketed (see Appendix \ref{appendix:Tokenization}). In the raw format, the grid is presented as a simple tokenized list without explicit spatial awareness. In contrast, the bracketed format introduces spatial awareness through the use of special bracket tokens that indicate the start and end of each row in the grid. These two input formats were then compared against models trained using the raw format combined with 2D positional encoding, referred to as 2D in our experiments, where spatial information is encoded explicitly at the positional encoding level rather than within the input.

Subsequently, we evaluated the sample efficiency of 2D positional encoding against more modern approaches, such as RoPE and Learned Embeddings, using a single model architecture to ensure a consistent and fair comparison.

\subsection{Different Architectures}
To evaluate the robustness of positional encoding strategies across varying model configurations, we investigated their performance in different architectural setups. Specifically, we examined the impact of positional encoding on:

\begin{itemize}
    \item \textbf{Model Size}: Small, Medium, and Large transformers (see Appendix \ref{appendix:Model size} for architecture details).
    \item \textbf{Decoder-Only Architecture}: A Medium model without encoder layers, focusing solely on the decoder’s capacity.
\end{itemize}

\begin{figure}[H]
    \centering
    \begin{subfigure}[t]{0.6\linewidth}
        \centering
        \includegraphics[width=\linewidth]{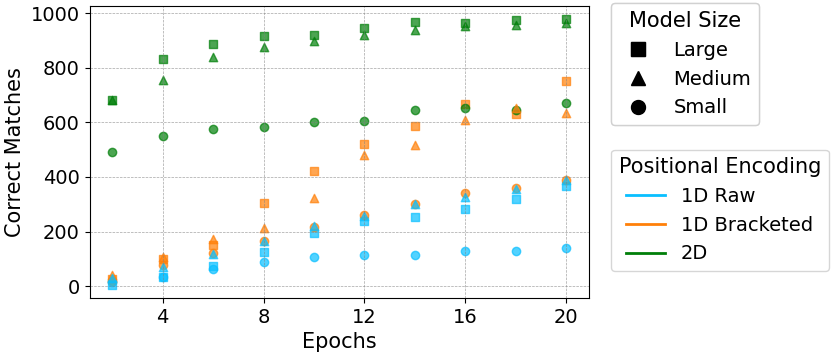}
        \caption{}
        \label{fig:across_size_performance}
    \end{subfigure}
    \hfill
    \begin{subfigure}[t]{0.6\linewidth}
        \centering
        \includegraphics[width=\linewidth]{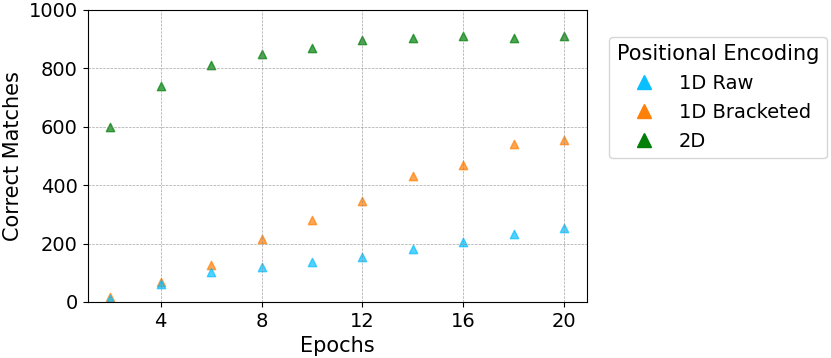}
        \caption{}
        \label{fig:decoder_only_performance}
    \end{subfigure}
    \caption{Impact of positional encoding: (a) shows the performance of 1D and 2D positional encodings across different model sizes, while (b) illustrates the performance of these encodings in a Medium decoder-only model configuration. The results highlight the consistent advantage of 2D positional encoding.}
    \label{fig:combined_impact_pe}
\end{figure}

Our results reveal a consistent trend: positional encoding plays a crucial role in determining model performance across all architectures (see Figure \ref{fig:combined_impact_pe}). Among the encoding strategies tested, 2D positional encoding stands out, demonstrating a clear advantage over 1D encoding, irrespective of model size or architectural configuration. This superiority holds even in models with a single transformer layer, where 2D encoding achieves competitive performance against larger models equipped with less effective positional encoding. Notably, even in a decoder-only architecture, where the model lacks the encoder's ability to view the entire grid, 2D encoding still performs remarkably well, achieving around 90\% accuracy.

\subsection{Modern approaches}
\begin{figure}[H]
    \centering
    \begin{subfigure}[t]{0.7\linewidth}
        \centering
        \includegraphics[width=\linewidth]{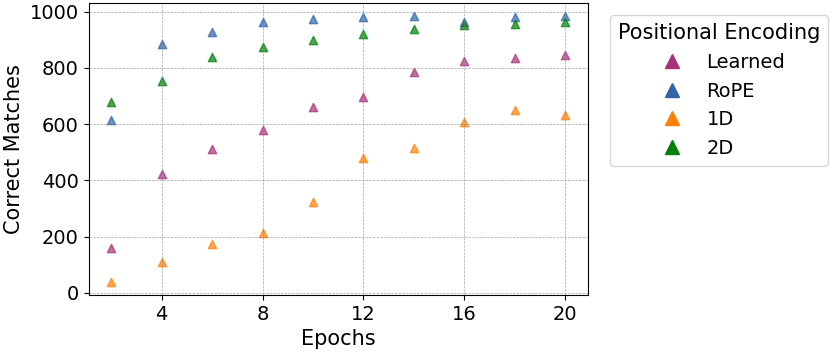}
        \caption{}
        \label{RoPE and Learned 100k}
    \end{subfigure}
    \hfill
    \begin{subfigure}[t]{0.7\linewidth}
        \centering
        \includegraphics[width=\linewidth]{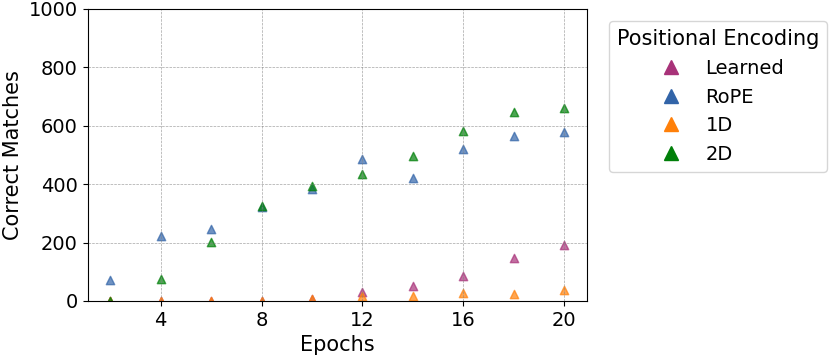}
        \caption{}
        \label{RoPE and Learned 10k}
    \end{subfigure}
    \caption{Impact of different positional encodings on a Medium model trained with (a) 100,000 examples and (b) 10,000 examples.}
    \label{RoPE and Learned}
\end{figure}

We provide a comparative analysis of several positional encoding strategies using a Medium model in two distinct scenarios: one with 100,000 training examples, where RoPE demonstrates a slight performance advantage over other methods (see Figure \ref{RoPE and Learned 100k}), and another with 10,000 training examples, where 2D encoding consistently achieves the best performance (see Figure \ref{RoPE and Learned 10k}). These results indicate that a clear and well-defined positional encoding has a direct impact on how effectively these models reason on ARC tasks. In particular, 2D positional encoding appears to be a more suitable choice given the limited data availability and the inability to generate additional examples during test time.

\section{Conclusion}

This work highlights the significant impact of positional encoding on the performance of transformer models in solving ARC tasks. Our findings suggest that while 2D positional encoding consistently outperforms other methods in data-constrained scenarios, it may not necessarily be the optimal choice if a robust dataset generator is available. However, it remains highly effective for training smaller models or achieving more efficient learning, making it a practical choice for scenarios with limited computational resources or data availability.

One limitation of this study is the treatment of ARC examples as independent, overlooking the reasoning often required between groups of related examples. Addressing this limitation could provide a deeper understanding of how positional encoding influences more complex relational reasoning. Future work should explore group-based reasoning while also expanding the analysis to include a broader range of tasks, building on the insights provided here to develop more effective positional encoding strategies for complex reasoning challenges.

\subsubsection*{Acknowledgments}
This work was supported by the project Center for Responsible AI reference no. C628696807-00454142, financed by the Recovery and Resilience Facility.

\bibliography{iclr2025_conference}
\bibliographystyle{iclr2025_conference}

\appendix
\section{Appendix} 
\subsection{Tokenization} 
\label{appendix:Tokenization}
\begin{figure}[H]
  \centering
\includegraphics[width=0.6\textwidth]{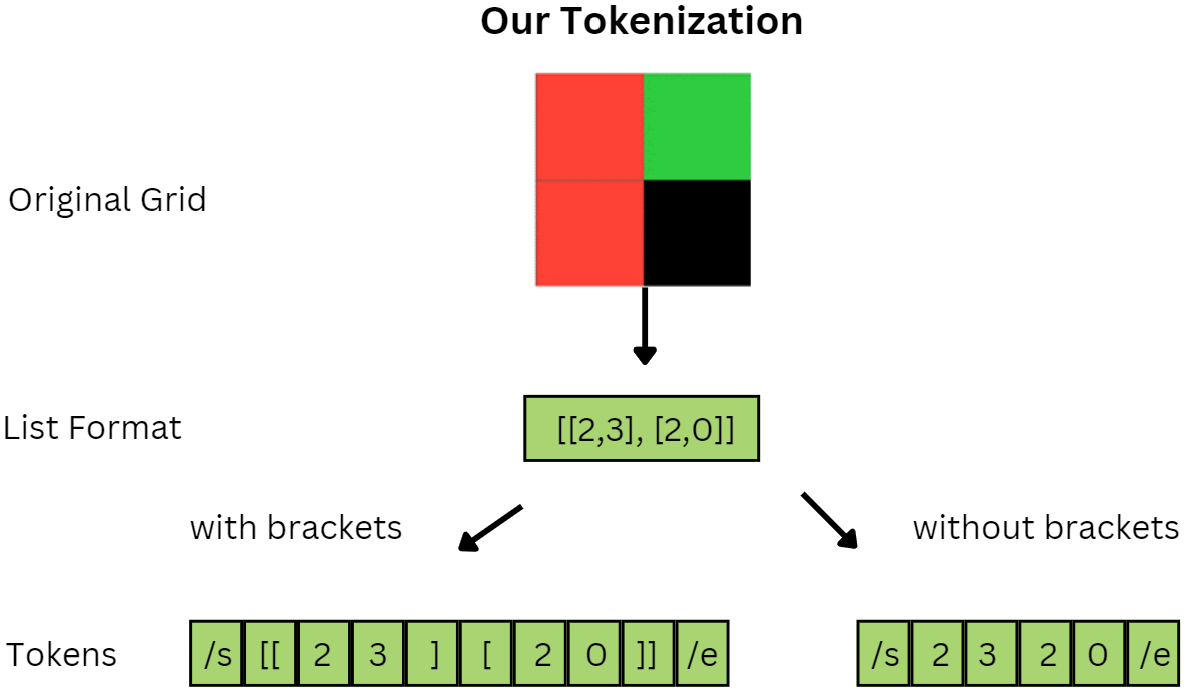}
     \caption{An illustration of the tokenization process used in the experiments. The original grid is first converted into a list format, where each number corresponds to a given color. In this case, 2 corresponds to red, 3 to green and 0 to black. The tokens are then generated in two ways: bracketed (left), where spatial structure is explicitly marked, and raw (right), where the grid is represented as a raw sequence of tokens. Both versions include start (/s) and end (/e) tokens to mark the boundaries of the sequence.}
  \label{Images/Our_Tokenizer}
\end{figure}

For these experiments, we used both raw and bracketed tokenization formats. Figure \ref{Images/Our_Tokenizer} illustrates how this mechanism works.

\subsection{Model size}
\label{appendix:Model size}
During this work, we used several transformer architectures to systematically evaluate the impact of positional encoding across different configurations. The specifications for these architectures, including Feedforward layer size, number of layers, and attention heads, are summarized in Table \ref{tab:transformer_configs}. Additionally, all models were trained using consistent hyperparameters, such as batch size, learning rate, and vocabulary size, as detailed in Table \ref{tab:consistent_specs}. These configurations were chosen to ensure a fair comparison while maintaining computational efficiency.  
 \begin{table}[H]
\caption{Difference between transformer architectures}
\label{tab:transformer_configs}
\centering
\begin{tabular}{@{}lccc@{}}
\toprule
\textbf{Model}             & \textbf{Feedforward Layer} & \textbf{Number of Layers} & \textbf{Number of Heads} \\ \midrule
Smaller Transformer        & 1024                      & 1                         & 4                        \\
Medium Transformer         & 1024                      & 4                         & 8                        \\
Larger Transformer         & 2024                      & 6                         & 8                        \\ \bottomrule
\end{tabular}
\end{table}

\begin{table}[H]
\caption{Consistent hyperparameters across all model architectures}
\label{tab:consistent_specs}
\centering
\begin{tabular}{@{}ll@{}}
\toprule
\textbf{Hyperparameter}         & \textbf{Value} \\ \midrule
Batch Size                      & 64             \\
Model Dimension                 & 512            \\
Dropout                         & 0.1            \\
Learning Rate                   & 0.0001         \\
Max Number of Tokens            & 512            \\
Seed Number                     & 42             \\
Vocabulary Size                 & 20             \\
Optimizer                       & AdamW \\
Loss                            & CrossEntropy \\ \bottomrule
\end{tabular}
\end{table}

\subsection{Datasets}
\label{appendix:Datasets}
For experiments involving CodeT5+, we trained exclusively on the task \texttt{22eb0ac0}, restricting the grid sizes to 15 by 15. Each model was trained on 1,000 examples generated for this task. 
 For the custom transformer models developed in this work, we utilized a broader set of tasks, including \texttt{22eb0ac0}, \texttt{36d67576}, \texttt{3aa6fb7a}, \texttt{08ed6ac7}, \texttt{e8593010}, \texttt{e21d9049}, \texttt{39e1d7f9}, \texttt{913fb3ed}, and \texttt{68b16354}. To manage memory and efficiency constraints, we restricted the grid sizes such that the sum of the dimensions of an example did not exceed 40.
 
\end{document}